\documentclass[letterpaper, 10 pt, conference]{ieeeconf}  

\IEEEoverridecommandlockouts                              

\makeatletter
\def\@maketitle{%
	\newpage
	\null
	\vskip -1.2cm   
	\begin{center} 
		{\LARGE \bfseries \@title \par}%
		\vskip 0.5em  
		{\normalsize \@author \par}%
		\vskip 0.5em
	\centering
	\includegraphics[width=0.95\linewidth
	]{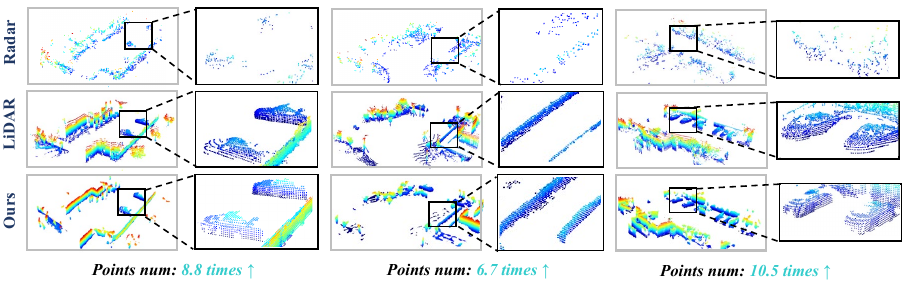}
	\captionof{figure}{Our method establishes a new state-of-the-art in sparse radar point cloud super-resolution, achieving a 6- to 10-fold increase in point cloud density while retaining key structural features of the scene, such as straight walls, curbs, and vehicles.}
	\label{fig::Absfig}
	\vspace{-20pt}  
	\end{center}%
	\par
	\vskip -0.1cm
}
\makeatother

\usepackage{amsmath}   %
\usepackage{amssymb}   %
\usepackage{caption}
\usepackage{float}
\usepackage{dblfloatfix}
\usepackage{booktabs}
\usepackage{graphicx}
\usepackage[pagebackref,breaklinks,colorlinks]{hyperref}

\title{\LARGE \bf
R2LDM: An Efficient 4D Radar Super-Resolution Framework Leveraging Diffusion Model
}
\author{Boyuan Zheng, Shouyi Lu$^\S$, Renbo Huang, Minqing Huang, Fan Lu, Wei Tian, Guirong Zhuo$^\dagger$  and Lu Xiong 
\thanks{*This work was supported by the National Natural Science Foundation of China under Grant 52325212, the National Key Research and Development Program of China under Grant 2022YFE0117100, the Fundamental Research Funds for the Central Universities, and Shanghai Tongyu Automobile Technology Intelligent Vehicle By-Wire Chassis Joint Laboratory. (Corresponding author: Guirong Zhuo)}
\thanks{$\dagger$ indicates the corresponding author: Guirong Zhuo (zhuoguirong@tongji.edu.cn)}
\thanks{$\S$ indicates the project leader: Shouyi Lu (2210803@tongji.edu.cn)}
\thanks{ The Authors are with School of Automotive Studies, Tongji University, Shanghai 201804, China.}
}

\begin{document}

\maketitle
\thispagestyle{empty}
\pagestyle{empty}

\begin{abstract}

We introduce R2LDM, an innovative approach for generating dense and accurate 4D radar point clouds, guided by corresponding LiDAR point clouds. Instead of utilizing range images or bird’s eye view (BEV) images, we represent both LiDAR and 4D radar point clouds using voxel features, which more effectively capture 3D shape information. Subsequently, we propose the Latent Voxel Diffusion Model (LVDM), which performs the diffusion process in the latent space. Additionally, a novel Latent Point Cloud Reconstruction (LPCR) module is utilized to reconstruct point clouds from high-dimensional latent voxel features. As a result, R2LDM effectively generates LiDAR-like point clouds from paired raw radar data. We evaluate our approach on two different datasets, and the experimental results demonstrate that our model achieves 6- to 10-fold densification of radar point clouds, outperforming state-of-the-art baselines in 4D radar point cloud super-resolution. Furthermore, the enhanced radar point clouds generated by our method significantly improve downstream tasks, achieving up to 31.7\% improvement in point cloud registration recall rate and 24.9\% improvement in object detection accuracy.

\end{abstract}

\section{INTRODUCTION}
\label{sec:intro}

4D millimeter-wave radar \cite{zhuoins20234drvo,zheng2022tj4dradset,ntuli20234d} has emerged as a critical sensor in autonomous driving, attracting significant attention from both academia and industry for its ability to provide 3D point cloud data and instantaneous velocity measurements. Additionally, the longer wavelength of 4D radar makes it nearly immune to interference from small particles in fog and rain, allowing it to operate reliably under challenging weather conditions. Compared to LiDAR and cameras, this robustness gives 4D radar a unique potential to support all-weather autonomous driving applications.

Despite these advantages, the point clouds provided by 4D Radar are two orders of magnitude lower than those from LiDAR \cite{zhang2024towards}, resulting in not only sparse data but also a lack of context information \cite{peng2024transloc4d}. In addition, multipath effects introduce artifacts, ghost points, and false targets into radar point clouds, making them susceptible to inaccuracies. Evidently, these factors significantly hinder the application of 4D radar in autonomous driving. This motivates us to seek an approach that effectively mitigates noise and sparsity of radar point clouds, enabling robust, continuous, and reliable all-weather sensing for downstream autonomous driving tasks.

It is noteworthy that recently, generative model approaches have achieved significant success in LiDAR point cloud upsampling and completion tasks, with diffusion models demonstrating the most outstanding performance. Given the powerful generative capabilities of diffusion models and their suitability for high-noise scenarios \cite{luan2024diffusion}, it is intuitive to consider adapting diffusion models to the task of radar point cloud super-resolution, using LiDAR as ground-truth supervision \cite{cheng2022novel,16_prabhakara2023high} .
However, directly applying diffusion to point clouds is challenging. Existing methods encode dense LiDAR data into a range image \cite{82_nakashima2024lidar} before the diffusion process. Given the fundamental differences in the point cloud generation principles of 4D radar and LiDAR, range image encoding is not well-suited for 4D radar point clouds, posing significant difficulties for this task.

To address these challenges, we propose a novel 4D radar super-resolution framework, R2LDM, which enhances sparse 4D radar points into dense, LiDAR-like points. Specifically, to develop a unified representation of both radar and LiDAR point clouds, we introduce the Latent Voxel Diffusion Model (LVDM), which encodes 4D radar and LiDAR point clouds as latent voxel features, performing the diffusion process in the latent space. We demonstrate that this approach captures point cloud features more effectively than range image encoding, achieving superior generative performance.

\setcounter{figure}{1}  
\begin{figure}[t]
	\centering
	\resizebox{1.0\columnwidth}{!}
	{
		\includegraphics[scale=1.00]{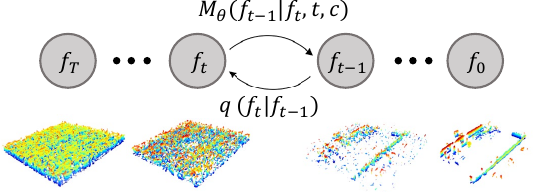}}
	\vspace{-5mm}
	\caption{ An illustration of our diffusion for point cloud super-resolution. During the forward process, we progressively add Gaussian noise on the ground truth LiDAR latent voxel feature $f_0$. A neural network $M_{\theta}$ is trained to denoise the noisy feature $f_T$ at time t based on condition c.}
	\label{sub:fig2}
	\vspace{-18pt}
\end{figure}

In contrast to similar work \cite{51_ran2024towards}, our diffusion model captures the distribution of point clouds at a high-dimensional feature level, thus requiring an additional module to reconstruct the point cloud back into real 3D space. Inspired by \cite{wang2022sparse2dense}, we propose a novel Latent Point Cloud Reconstruction (LPCR) module, which employs sparse convolution to predict an occupancy probability mask and offset for each voxel from latent features. Experimental results demonstrate that this approach robustly predicts voxel points closely approximating the ground truth.

To alleviate the training challenges caused by the large difference in point cloud density between the two modalities (e.g., the presence of more empty voxels in Radar data), we divide the training process into two stages. In the first stage, we train the LPCR module to enable the model to reconstruct 3D point clouds from sparse features in the latent space, and in the second stage, we train the LVDM module to learn the feature distribution from 4D Radar to LiDAR. We conducted experiments on both public and self-made datasets, demonstrating that R2LDM outperforms all baseline methods in the point cloud super-resolution task. Additionally, we validated the effectiveness of the generated point cloud in downstream tasks such as point cloud registration. Overall, the contributions of this work are summarized as follows:

\vspace{0.1cm}
\begin{itemize}
	\item We propose a novel 4D mmWave radar point clouds super-resolution method that generates dense and accurate LiDAR-like point clouds. To the best of our knowledge, this is the first approach to model the transformation from 4D Radar to LiDAR utilizing a latent space diffusion process.
	
\end{itemize}
\vspace{-0.1cm}
\begin{itemize}
	\item We propose an innovative Latent Point Cloud Reconstruction (LPCR) module that directly reconstructs point clouds from latent features. This approach ensures robust voxel-wise predictions and achieves reconstruction results that closely match the ground truth.
	
\end{itemize}
\vspace{-0.1cm}
\begin{itemize}
	\item We validate our framework on two different datasets, demonstrating that our method outperforms state-of-the-art baselines, achieving up to 31.7\% improvement in point cloud registration recall rate and 24.9\% improvement in object detection accuracy compared to the original radar point cloud.
\end{itemize}
\vspace{-0.1cm}

\section{Related Work}
\label{sec:formatting}

\begin{figure*}[t]
	\centering
	\includegraphics[width=0.9\linewidth]{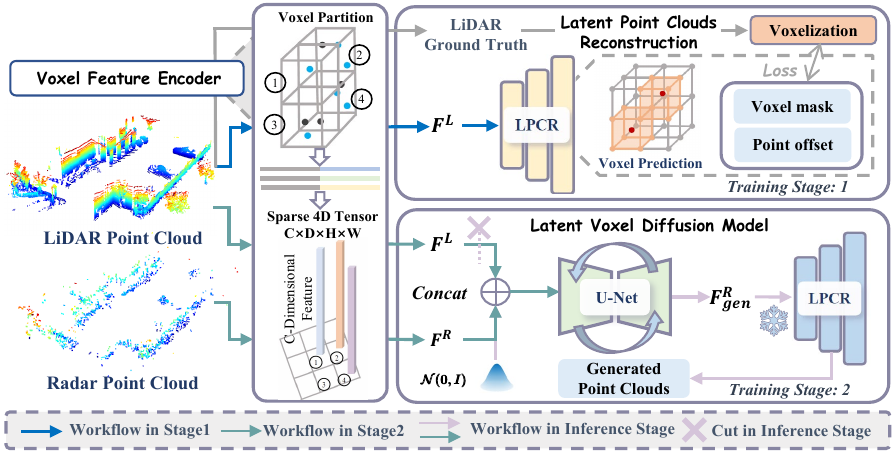}
	\caption{
		An overview of our R2LDM framework. Our training is divided into two stages: In Stage1, LiDAR point clouds are encoded into latent voxel features using the VFE module, and the LPCR module learns to reconstruct 3D point clouds from these latent voxel features. In Stage 2, the LVDM module takes the features of both LiDAR and 4D radar point clouds as input, learning to approximate radar features to LiDAR features in the latent space. During the inference phase, the model takes Gaussian noise and raw radar point clouds as input and accurately generates dense LiDAR-like point clouds, resulting in up to 10 times densification.}
	\label{pic:main}
	\vspace{-0.5cm}
\end{figure*}

\noindent\textbf{Radar Point Cloud Super-Resolution} Typically, sparse and noisy radar point clouds can be obtained using conventional target detectors such as CFAR \cite{choset2005principles} and MUSIC  \cite{schmidt1986multiple}. Compared to post-processing techniques for radar point clouds \cite{zhang20234dradarslam,4_huang2024multi}, learning-based point cloud densification methods exhibit superior potential. Chamseddine \cite{3_chamseddine2021ghost} employs PointNet \cite{17_charles2017pointnet} to differentiate between ghost and real targets, leading to precise radar point cloud generation. Prabhakara \cite{16_prabhakara2023high} introduces RadarHD, which utilizes a U-Net \cite{19_ronneberger2015u} to transform low-resolution radar point clouds into dense LiDAR-like point clouds. Zhang \cite{zhang2024towards} proposed a single-chip mmWave radar point cloud construction method designed for autonomous navigation of MAVs.  However, the aforementioned works are primarily based on 2D data. In contrast, our R2LDM enables effective 3D point cloud densification while preserving the scene context.


\noindent\textbf{Difffusion Models} Recently, generative models \cite{song2020improved,ho2020denoising,saharia2022photorealistic,goodfellow2020generative} have garnered widespread attention and found broad applications. Among these, diffusion models \cite{sohl2015deep} have achieved significant success in the field of image generation with notable advancements such as LDMs \cite{rombach2022high}.
Researchers have also developed language-guided diffusion models \cite{nichol2021glide,ramesh2022hierarchical}  and other controllable diffusion models \cite{zhang2023adding}, which demonstrate the immense potential of diffusion models. Our R2LDM is also a conditional diffusion model, capable of taking sparse millimeter-wave radar point clouds as conditional inputs and generating dense LiDAR-like point clouds as output.

\noindent\textbf{LiDAR Scene Generation} Generative models offer a promising approach for synthesizing realistic LiDAR point clouds without the need to reconstruct real-world environments. Initial efforts leveraged range image representation to facilitate LiDAR synthesis. The pioneering work by Caccia \cite{8_caccia2019deep} explore the application of generative models to LiDAR scenes, proposing LiDARVAE and LiDARGAN for unconditional generation and noisy LiDAR reconstruction. R2DM \cite{82_nakashima2024lidar} presented a novel generative model for LiDAR data that can generate diverse and high-fidelity 3D scene point clouds based on the image representation of range and reflectance intensity. To make the generation process more controllable, researchers try to let DMs support conditioning on semantic layouts or text. LiDM \cite{51_ran2024towards} proposed a generative model that consumes arbitrary input conditions (e.g., layouts, text) for LiDAR-realistic scene generation.

Luan \cite{luan2024diffusion} introduces a mean-reverting SDE-based diffusion model for 4D mmWave radar point cloud super-resolution. However, this approach converts radar point clouds into BEV images, where the height is encoded in the channel. This scheme leads to each pixel representing at most one point, resulting in sparse point clouds and limiting performance. In contrast, our work encodes the point cloud as latent voxel features, performs the diffusion process in the latent space, and effectively densifies the radar point cloud through the LPCR module, significantly enhancing the point cloud resolution. The visualization results are presented in Fig.~\ref{fig::Absfig}.

\section{Proposed Method}

This section expounds on the details of the R2LDM.
In Sec.~\ref{sec:represent}, we discuss our design choice of latent voxel features for data representation.
In Sec.~\ref{sec:problem}, we formulate the task of enhanced radar point clouds generation with R2LDM.
In Sec.~\ref{sec:rcd}, we define the latent voxel diffusion model with 4D radar as the condition.
In Sec.~\ref{sec:pcr}, we present how the latent point clouds reconstruction module work.
In Sec.~\ref{sec:train}, we define the training objectives in two stages.
Fig.~\ref{pic:main} illustrates the overall R2LDM framework.

\subsection{Data Representation} 
\label{sec:represent}
\begin{figure}[t]
	\centering
	\resizebox{1.0\columnwidth}{!}
	{
	\includegraphics[scale=1.00]{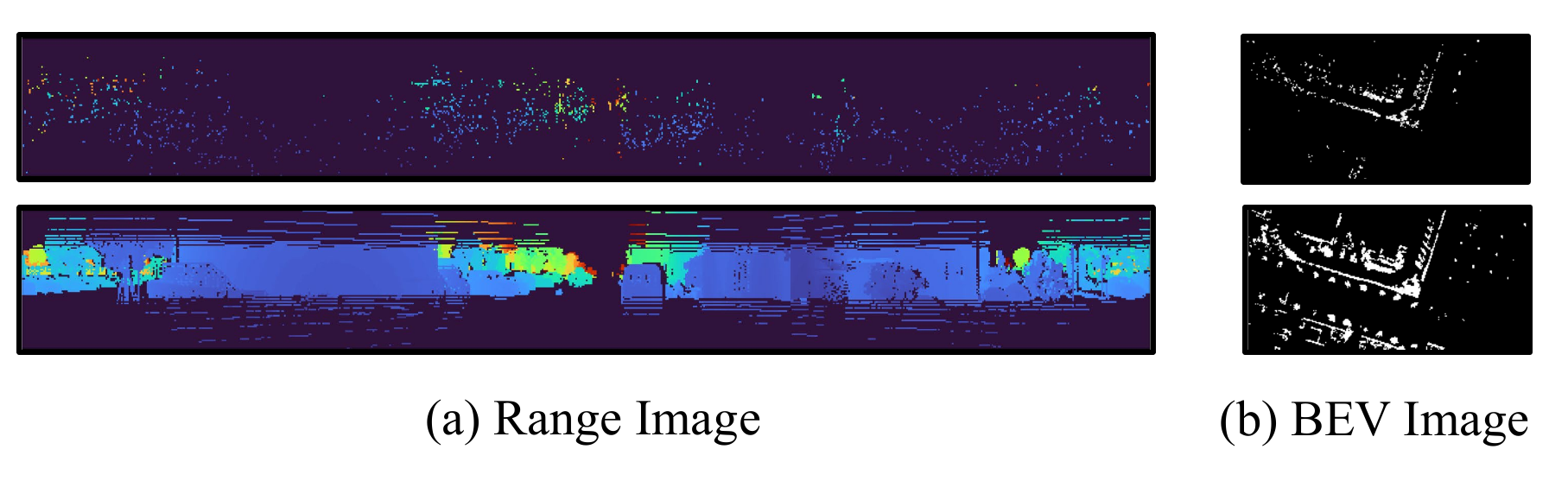}}
	\vspace{-5mm}
	\caption{(a) Range images preserve more point information but often result in inaccurate reconstructions. (b) BEV images lose crucial height information, making both methods inadequate for our super-resolution task.}
	\label{fig::bev}
	\vspace{-15pt}
\end{figure}

Effective representation of point cloud data is crucial for radar-to-LiDAR super-resolution. Existing methods, such as range image and BEV image encoding, struggle to adapt to both radar and LiDAR point clouds due to their distinct generation processes. As shown in Fig.~\ref{fig::bev}, these differences hinder the use of traditional encoding methods for both point cloud types. To address this, we adopt a voxel feature encoder, which better captures point cloud data and preserves high-dimensional latent features, facilitating the diffusion process. More details on data processing are in Sec.~\ref{sec:dataprocess}.

\subsection{Problem Formulation}
\label{sec:problem}

The input point clouds and our generated point clouds are denoted as $P$ and $\hat{P}$, respectively, where $P, \hat{P}\in\mathbb{R}^{N \times 3}$.  As described in Sec.~\ref{sec:represent} , we use the Voxel Feature Encoder $\mathcal{E}$ to encode $P$  into latent feature $F=\mathcal{E}(P)$, where $F\in\mathbb{R}^{h \times w \times d}$. Our Latent Point Cloud Reconstruction module serves as a decoder $\mathcal{D}$ to reconstruct $\hat{P}=\mathcal{D}(F)=\mathcal{D}(\mathcal{E}(P))$. For the diffusion process, we formulate the point cloud voxel feature as the latent variable, and LiDAR-like latent feature are generated iteratively from the reverse diffusion process.

\subsection{Latent Voxel Diffusion Model}
\label{sec:rcd}

Directly recovering a densified point cloud feature from pure noise is non-trivial. Therefore, we condition the process on radar features $F^{R}$ and approximate the distribution of the high-dimensional features $F^{R}$ and $F^{L}$ in the latent space, thereby controlling the diffusion generation process.
\textbf{Forward diffusion process} While training, the forward diffusion process gradually adds Gaussian noise into the ground truth LiDAR latent feature $f_0$ for $T$ timesteps via a Markov chain:

\vspace{-2mm}
\begin{equation}
	\label{eq:feature gathering}
	q(f_{1:T}|f_{0}) = \Pi_{t=1}^{T}  q(f_{t}|f_{t-1}),
	\vspace{-1mm}
\end{equation}
where $f_{0}$ indicates the ground truth latent feature. $f_{t}$ is the intermediate feature at timestamp $t$. Gaussian transition kernel $q(f_{t}|f_{t-1}) =\mathcal{N} (f_{t}; \sqrt{1-\beta_{t}} f_{t-1}, \beta_{t}I)$ is fixed by predefined hyper-parameters $\beta_{t}$.The forward process progressively adds noise to the ground truth feature $f_{0}$ and eventually turns it into a total Gaussian noise when $T$ is large enough.

\textbf{Reverse denoising process} Basically, the reverse process can be represented as a parameterized Markov chain starting from a random noise $p(f_{T})$:
\vspace{-2mm}
\begin{equation}
	\label{eq:feature gathering}
	p_{\theta}(f_{0:T}) = p(f_{T})\Pi_{t=1}^{T} p_{\theta}(f_{t-1}|f_{t}),
	\vspace{-2mm}
\end{equation}
where the reverse transition kernel $p_{\theta}(f_{t-1}|f_{t})$ is approximated with a neural network. We directly learn the latent variable $\hat{f}_{0}$ by the denoising network $M_{\theta}(f_{t-1}|f_{t},t)$.

\textbf{Design of condition signals}

Specifically, in our task, a set of spatially and temporally aligned LiDAR features \textbf{${f}_0$} and radar feature \textbf{$c$} is provided through a voxel feature encoder. Condition signals $c\in\mathbb{R}^{h \times w \times d}$ guide the reverse process to progressively produce the densified point cloud feature $\hat{f}_{0} = M_{\theta}(f_{t},t,c)$.

\textbf{Inference phase} LVDM takes pure Gaussian noise \textbf{$z$} as input, conditioned on a latent voxel radar feature \textit{c}, to recover the original data \textbf{${{f}_0}$} and reconstruct the point cloud $\hat{P}$ through the LPCR module. The visualization results are presented in Fig.~\ref{sub:fig2}.

Similar to \cite{82_nakashima2024lidar}, our model is built upon Efficient U-Net. Compared to image-based diffusion models \cite{8_caccia2019deep,51_ran2024towards}, our LVDM is more suitable for likelihood-based generative models, as it enables a focus on key semantic information within a lower-dimensional and computationally efficient space.

\subsection{Latent Point Cloud Reconstruction Module}
\label{sec:pcr}
Considering the challenges of directly reconstructing large-scale dense scene points \cite{xu2021spg}, we adopt a voxel-level reconstruction approach inspired by Sparse2Dense \cite{wang2022sparse2dense}, predicting only the average position of input points within each non-empty voxel. By redesigning the prediction module and loss function, our method robustly predicts the entire scene’s point clouds that closely approximate the ground truth.

\begin{figure}[t]
	\centering
	\resizebox{1.0\columnwidth}{!}
	{
	\includegraphics[scale=0.85]{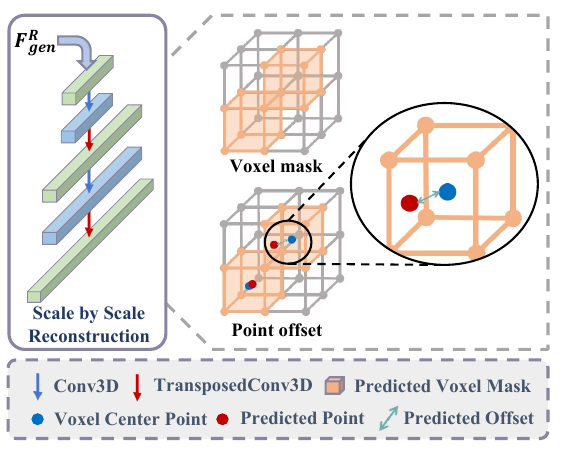}}
	\vspace{-5mm}
	\caption{The arcitecure of LPCR moduel. The features $F_{gen}^R$ are processed through Conv3D and TransposedConv3D modules to predict voxel masks $\mathcal{\mathcal{M}}_{\textit{mask}}$ and point cloud offsets $\mathcal{\mathcal{B}}_{\textit{offset}}$ at different scales, enhancing predictions robustness.}
	\label{fig::LPCR}
	\vspace{-16pt}
\end{figure}

Specifically, the LPCR module performs two key tasks: first, it predicts a soft voxel-occupancy mask that estimates the likelihood of each voxel being non-empty; second, it predicts a point offset $\mathcal{\mathcal{B}}_{\textit{offset}}$ for each non-empty voxel, which represents the displacement from the voxel center $V_{center}$ to the averaged input points within that voxel. Fig.~\ref{fig::LPCR} illustrates the architecture of the LPCR module.
The densified feature $F^{L}$ is first re-projected into a 3D space and processed through a 3D convolutional layer. A transposed 3D convolution upscales the output to one-quarter of the original resolution. A 1×1 3D convolution with sigmoid activation function predicts the voxel-occupancy mask $\mathcal{\mathcal{M}}_{\textit{mask}}$ , while another 1×1 3D convolution predicts the point offset $\mathcal{\mathcal{B}}_{\textit{offset}}$. This process is repeated at half and full scales to predict $\mathcal{\mathcal{M}}^{i}_{\textit{mask}}$ and $\mathcal{\mathcal{B}}^{i}_{\textit{offset}}$ at each scale $(i = \frac{1}{4}, \frac{1}{2}, 1)$. The final reconstructed point cloud $P_{rec}$ is calculated as:
\vspace{-2mm}
\begin{eqnarray}
	\label{eqnarray_1}
	P_{rec} = \left (  V_{center}  + \mathcal{\mathcal{B}}_{\textit{offset}} \right )  \times  \mathcal{\mathcal{M}}_{\textit{mask}},
	\vspace{-4mm}
\end{eqnarray}
where $P_{rec}$ is the reconstructed dense point cloud, and all variables in (\ref{eqnarray_1}) are considered at the scale $i = 1$.

The restoration of 3D points from high-dimensional features $F^L$ on a scale-by-scale basis enhances the robustness of the process, making it better suited for our specific task. During training, the LiDAR point cloud is used as the ground truth and is voxelized for supervised learning.

\vspace{-2mm}
\subsection{Training Process}
\label{sec:train}
\subsubsection{Training Strategy}
As illustrated in Fig.~\ref{pic:main}, our training is divided into two stages. Due to the significant disparity in point cloud density between LiDAR and 4D radar, directly training all modules together would present significant challenges. To address this, we adopt a strategy similar to that in \cite{rombach2022high}, where we first train the Voxel Feature Encoder (VFE) and LPCR modules, followed by the training of the diffusion model. In Stage1, $P^{L}$ is encoded into a latent voxel feature $F^{L}$ through the VFE module, which is then input into the LPCR module to learn dense point cloud reconstruction from $F^{L}$. In Stage2, LiDAR and radar point clouds are encoded using separate encoders, with the LiDAR encoder being  frozen. The radar encoder is then jointly trained with LVDM, enabling the conditional diffusion model to learn approximate radar features $F^{R}$ and LiDAR features $F^{L}$ in the latent space. The results demonstrate that this two-stage training approach allows the model to effectively learn key representations while conserving computational resources and promoting faster convergence.

\begin{figure}[t]
	\centering
	\resizebox{0.8\columnwidth}{!}
	{
	\includegraphics[scale=1.00]{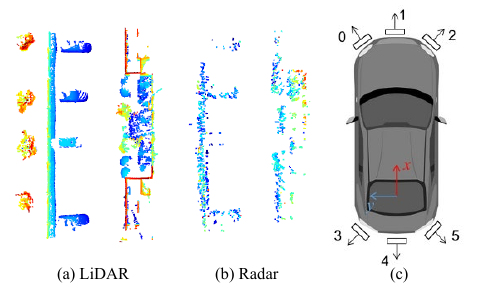}}
	\vspace{-3mm}
	\caption{LiDAR and 4D radar point clouds from the same scene in GPAL dataset. 
		(a) LiDAR: Generated from a RS-Ruby-Plus LiDAR, producing 24,713 points in this frame.
		(b) 4D Radar: Generated from six 4D radars, yielding 1,865 points in this frame.
		(c) Radar Sensor Configuration: The arrangement of the six radars enables full 360° FOV coverage.}
	\label{fig::supfig1}
	\vspace{-16pt}
\end{figure}

\subsubsection{Training Objectives} 

\noindent\textbf{LVDM}~
Diffusion Models (DMs) aim to comprehend a data distribution through a progressive denoising process on variables sampled from a Gaussian distribution. In our work, we enable DMs to leverage a lowdimensional latent space created by our LVFE module. This space adequately preserves LiDAR point clouds of scenes while maintaining computational efficiency. Given an input condition $c$, we define the objective of our LVDM:
\vspace{-2mm}
\begin{equation}
	\mathcal{\mathcal{L}}_{\textit{stage2}}=\left\|f_0 - M_\theta\left(f_t, t,c\right)\right\|_2^2,
\end{equation}
where $M_\theta\left(f_t, t,c\right)$ is a UNet backbone with timestamp conditioning.

\noindent\textbf{LPCR}~Multi-scale 3D convolutions (conv3d) are used to predict the voxel occupancy mask $\mathcal{\mathcal{M}}_{\textit{mask}}$ and point offset $\mathcal{\mathcal{B}}_{\textit{offset}}$ at different scales. The features generated by the convolutions are transformed into values between 0 and 1 via a softmax operation, then compared with the ground truth using binary cross-entropy loss, defined as: 
\vspace{-2mm}
\begin{equation}
	\operatorname{\mathcal{L}^{i}_{\textit{mask}}}=-\frac{1}{N} \sum_{i=1}^{N}\left[v_{i} \cdot \log \left(\hat{v}_{i}\right)+\left(1-v_{i}\right) \cdot \log \left(1-\hat{v}_{i}\right)\right],
	\vspace{-1mm}
\end{equation}
where $v_{i}$ denotes an element from the ground truth mask, and $\hat{v}_{i}$ represents the corresponding element from the generated mask. 
The offset feature $\mathcal{\mathcal{B}}_{\textit{offset}}$ is transformed into a positive value $\mathcal{\mathcal{B}}^{'}_{\textit{offset}}$ within the range (0,1) using a sigmoid function. We set the maximum offset value in each direction to half the length of the largest side of the voxel grid, denoted as $l_{length}$, thereby constraining the offset within each voxel to (0, $l_{length}$) in all directions. We then compute the $L_{1}$ loss between the true offset and the predicted offset, defined as:
\vspace{-2mm}
\begin{eqnarray}
	\mathcal{\mathcal{L}}^{i}_{\textit{offset}}=\frac{1}{N} \sum_{i=1}^{N}\left|\mathbf{\textit{gen}}_{i}-\mathbf{\textit{gt}}_{i}\right|,
	\vspace{-3mm}
\end{eqnarray}
where $gen_i$ represents the generated offset value, and $gt_i$ corresponds to the ground truth. 
Thus, in training Stage1, the full training loss of the PCR module is then written as:
\vspace{-2mm}
\begin{eqnarray}
	\mathcal{\mathcal{L}}_{\textit{stage1}} = \lambda _{1}\sum_{i=1}^{N}\mathcal{\mathcal{L}}_{\textit{mask}} + \lambda _{2}\sum_{i=1}^{N}\mathcal{\mathcal{L}}_{\textit{offset}},
	\vspace{-3mm}
\end{eqnarray}
where $\lambda_1$ = 0.9 and $\lambda_2$ = 0.1 are weights for each loss function.

\section{Experiments}
\label{sec:formatting}
\subsection{Benchmark Datasets}

\noindent\textbf{View-of-Delft(VoD) Dataset} The VoD dataset \cite{palffy2022multi} contains 8,682 frames of data collected by a Velodyne HDL-64 LiDAR and a ZF FRGen21 4D radar in complex urban traffic environments. We adopt the setup of Lu \cite{lubo},  utilizing 7,539 frames from 22 sequences for training and 1,143 frames from 5 sequences (04,09,16,20,23) for testing. The test set contains frames in unseen sequences, allowing for a comprehensive evaluation of the generalization ability.

\noindent\textbf{Geometry-PAL(GPAL) Dataset} The GPAL dataset contains 18708 frames of data collected by a RoboSense RS-Ruby-Plus LiDAR and six GPAL-Ares-R7861 4D radars in complex urban park environments in 8 different scenes. We divide it into 16,632 frames for training and 2,076 frames for testing. As illustrated in Fig.~\ref{fig::supfig1}, six 4D Radars are positioned around the vehicle, providing the radar system with a 360° field of view (FOV) that matches that of the LiDAR.

\vspace{-2mm}
\subsection{Data Processing}
\label{sec:dataprocess}
\noindent\textbf{Ground Point Removal} Ground points are first removed from the raw point cloud data, as they offer limited semantic value and can hinder the super-resolution learning process. Due to the low resolution of radar echo intensity, radar point clouds inherently contain few ground points, thus requiring no additional ground removal. For LiDAR data, we apply Patchwork++ \cite{lee2022patchwork++} to detect and exclude ground points. Patchwork++ employs adaptive ground likelihood estimation to iteratively approximate ground segmentation areas, achieving accurate ground separation even in complex terrains with multiple elevation layers.

\noindent\textbf{Shared field of view alignment} To represent point clouds using voxel features, we first restrict the height of both radar and LiDAR point clouds to the same range and focus on their overlapping horizontal field of view (FOV). In the VoD dataset,we consider the LiDAR with a $360^{\circ}$ horizontal FOV and radar with a $120^{\circ}$ horizontal FOV, retaining points with yaw angles within $\theta \in \left [ 30^{\circ}, 150^{\circ} \right ]$.
For the GPAL dataset, with a LiDAR and six 4D radars providing a combined $360^{\circ}$ horizontal FOV, we align the points based on their shared field of view.

\noindent\textbf{Voxel Generation} The LiDAR and radar point clouds are converted into voxel grids while maintaining the same spatial range for all points. For the GPAL dataset, with aligned radar and LiDAR FOVs, we retain points in $x\in [-16, 16]$, $y\in\left [ -16 , 16  \right ] $, and $z\in\left [ -0.5, 3.5  \right ] $ (in meters), and compress them into a 256 × 256 × 40 spatial voxel grid with a resolution of 0.125 m. For the VoD dataset, which has a single front-facing 4D radar, we select points with $x> 0$, retaining $\left [ 0 , 16  \right ] $, $y\in\left [ -8 , 8  \right ] $, and $z\in\left [ -0.5, 3.5  \right ] $ (in meters), with a voxel grid resolution of 6.25 cm.

\noindent\textbf{Multi-Frame Input} Due to the sparsity of radar point clouds, we aggregate data from multiple consecutive radar frames by aligning them based on their relative poses. In the VoD dataset, we use radar point clouds from 5 consecutive frames as input to the network, while in the GPAL dataset, we use point clouds aggregated from six temporally synchronized radar frames.

\noindent\textbf{Implementation Detail}  Our model has 40.1M parameters in total. We train our model on a single NVIDIA RTX 2080Ti with a batch size of 1, using the AdamW optimizer. In Stage1 of training, the initial learning rate is set to $1\times 10^{-3} $  , with a training duration of 9 hours.  In Stage2 of training, the initial learning rate is set to $1\times 10^{-4} $, and the training duration is 21 hours.

\begin{figure}[t]
	\centering
	\resizebox{0.96\linewidth}{!}
	{

	\includegraphics[scale=0.80]{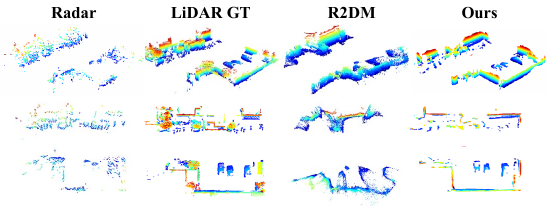}}
\vspace{-2mm}
\caption{Qualitative results on GPAL dataset.}
\label{fig4}
\vspace{-13pt}
\end{figure}

\begin{figure}[t]
	\centering
	\resizebox{1.0\linewidth}{!}
	{

	\includegraphics[scale=1.0]{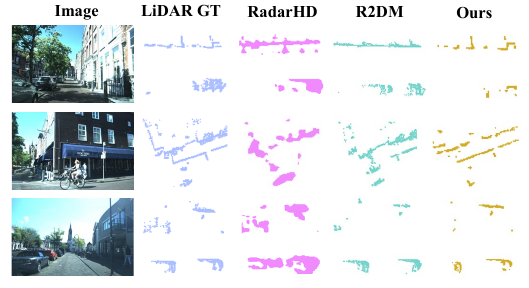}}

\vspace{-3mm}
\caption{Qualitative comparisons of 2D point clouds conducted on VoD dataset.}
\label{sub:fig3}
\vspace{-20pt}
\end{figure}

\begin{table}[H]
	\centering
	\huge 
	\resizebox{1.00\linewidth}{!}{ 
		\begin{tabular}{@{}l c  @{\hskip 10pt}c cccc}
			\toprule
			Method& CD $\downarrow$& HD $\downarrow$& F-Score $\uparrow$& FPD $\downarrow$& JSD\textsubscript{BEV} $\downarrow$&MMD\textsubscript{BEV} $\times 10^{-3} $$\downarrow$\\
			\midrule
			(a) GPAL Dataset& & & & & &\\
			R2DM\cite{82_nakashima2024lidar}& 29.05& 91.31& 0.10& 5.03& 0.32&29.9
			\\
			Ours & $\textbf{0.79}$ &$\textbf{24.17}$ &$\textbf{0.48}$ &$\textbf{3.85}$ &$\textbf{0.09}$ &$\textbf{0.86}$\\
			\bottomrule
			(b) VoD Dataset& & & & & &\\
			R2DM\cite{82_nakashima2024lidar}& 7.36& 23.63& 0.08& 4.89& 0.46& 22.6\\
			Ours & $\textbf{0.64}$& $\textbf{16.16}$& $\textbf{0.43}$& $\textbf{1.52}$& $\textbf{0.39}$& $\textbf{1.20}$\\
			\bottomrule
			& & & & & &\\
		\end{tabular}
	}
	\vspace{-4mm}
	\caption{Quantitative comparisons on both datasets.}
	\label{tab:Quantitative}
	\vspace{-6mm}
\end{table}

\subsection{Performance on Point Cloud Super-Resolution}
\label{sec:pcsuper}
\subsubsection{3D Metric}  
Our evaluation metrics include CD, HD, F-Score, FPD, JSD, and MMD, commonly used in previous works \cite{82_nakashima2024lidar,zhang2024towards}.

\begin{table}[t]
	\centering
	\renewcommand{\arraystretch}{1.6}
	\setlength\tabcolsep{2pt}
	\resizebox{1.0\linewidth}{!}{
		\fontsize{9}{9}\selectfont
		\begin{tabular}{l|ccccc}
			\hline
			\specialrule{0em}{2pt}{0pt}
			\makebox[0.1\textwidth][l]{Methods} &  \makebox[0.07\textwidth]{CD \scalebox{1}{$\downarrow$}}  & \makebox[0.07\textwidth]{MHD \scalebox{1}{$\downarrow$}}  &  \makebox[0.07\textwidth]{UCD \scalebox{1}{$\downarrow$}}  & \makebox[0.07\textwidth]{UMHD \scalebox{1}{$\downarrow$}} & \makebox[0.07\textwidth]{FID$_\text{BEV}$ \scalebox{1}{$\downarrow$}}  \\
			
			\specialrule{0em}{0pt}{2pt}
			\hline
			\specialrule{0em}{2pt}{0pt}
			
			RadarHD\cite{16_prabhakara2023high} & 0.34 & 0.30 & 0.15 & 0.10 & 243.57\\
			R2DM\cite{82_nakashima2024lidar} & 0.23 & 0.17 & 0.17  & 0.06 & 202.91\\ 
			Luan\cite{luan2024diffusion} & $\textbf{0.19}$ & $\textbf{0.10}$ & 0.15  & 0.07 & $\textbf{118.40}$\\ 
			Ours & $\textbf{0.19}$ & 0.12 & $\textbf{0.14}$ & $\textbf{0.04}$ & 161.17\\ 
			
			\specialrule{0em}{2pt}{0pt}
			
			\hline
		\end{tabular}%
	}
	\captionsetup{justification=centering} 
	
	\caption{Quantitative comparisons on VoD dataset.}
	\label{tab:quantative_self_1}
	\vspace{-3mm}
\end{table}

\begin{table}[t]
	\centering
	\resizebox{0.86\linewidth}{!}{ 
		\begin{tabular}{c  @{\hskip 10pt}c cc}
			\toprule
			& & $[succ./all]$&$[succ./all]$\\
			Method& RR(\%) $\uparrow$& RE($^\circ$) $\downarrow$&TE(m)$\downarrow$\\
			\midrule
			Raw& 85.7& 0.79/$\textbf{0.84}$& 0.18/0.37\\
			R2DM\cite{82_nakashima2024lidar}& 77.8& 0.79/1.21& 0.25/0.41\\
			Ours& $\textbf{89.0}$& $\textbf{0.74/}$0.86& $\textbf{0.17/0.28}$\\
			\bottomrule
		\end{tabular}
	}
	\caption{Registration performance on VoD dataset.}
	\label{tab:Regis}
	\vspace{-20pt}
\end{table}

\begin{table}[ht]
	\begin{center}
		\resizebox{1.00\linewidth}{!}
		{
			\begin{tabular}{lcccc|cccc}
				\hline
				\toprule
				&\multicolumn{4}{c|}{Entire Annotated Area }&\multicolumn{4}{c}{In Driving Corridor }\\
				Points & Car & Pedestrian & Cyclist & mAP & Car & Pedestrian & Cyclist & mAP\\ \midrule	
				Raw   & 33.07 &18.54 & 23.15& 24.92 &70.59 &27.33 & 44.32 & 47.41 \\
				Ours  
				&\bf 33.11 & \bf 23.89
				&\bf 34.10 & \bf 30.37	
				&\bf 71.18 & \bf 34.76
				& \bf 69.20 & \bf 58.38\\
				\bottomrule
				\hline
			\end{tabular}
		}
		\vspace{-5pt}
		\caption{Object detection results on VoD dataset.}
		\label{table:flyingthing3d}
	\end{center}
	\vspace{-8mm}
\end{table}

\begin{figure}[t]
	\centering
	\resizebox{0.8\linewidth}{!}
	{
		\includegraphics[scale=1.0]{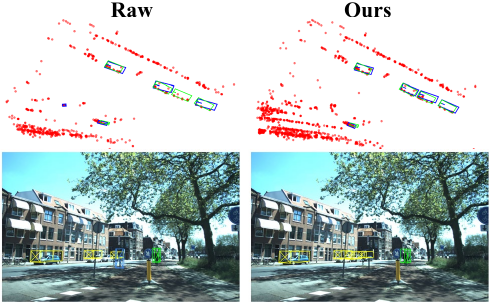}}
	\vspace{-2mm}
	\caption{Object detection results on the VoD dataset. Note that in the first row, blue boxes show the detection results and green boxes show the ground truths. In the second row, we present the detection results of car (yellow), cyclist (green), and pedestrians (blue), with R2LDM enhanced point cloud detection having a higher accuracy rate.}
	\label{sub:fig::od}
	\vspace{-22pt}
\end{figure}

\begin{table}[t]
	\centering
	\huge 
	\resizebox{1.00\linewidth}{!}{ 
		\begin{tabular}{@{}l c  @{\hskip 10pt}c cccc}
			\toprule
			Method& CD $\downarrow$& HD $\downarrow$& F-Score $\uparrow$& FPD $\downarrow$& JSD\textsubscript{BEV} $\downarrow$&MMD\textsubscript{BEV} $\times 10^{-3} $$\downarrow$\\
			\midrule
			(a) Diffusion& & & & & &\\
			w/o diffusion& 19.13& 168.78& 0.19& 193.07& 0.35&24.80\\
			with MLP-based $M_{\theta }$&8.36&52.83&0.20&18.67&0.25&4.40\\
			with transformer-based $M_{\theta }$& 3.69& 54.67& 0.21& 36.97& 0.13&2.20\\
			Ours(UNet-based $M_{\theta }$ )& $\textbf{0.79}$& $\textbf{24.17}$& $\textbf{0.48}$& $\textbf{3.85}$& $\textbf{0.09}$&$\textbf{0.86}$\\
			\bottomrule
			(b) Condition& & & & & &\\
			Attention& 9.65& 61.77& 0.25& 49.66& 0.31&6.73\\
			with geometry feature& 8.77& 71.98& 0.11& 9.05& 0.28&7.92\\
			with intensity/power& 0.82& $\textbf{23.91}$& 0.47& 4.08& $\textbf{0.09}$&0.90\\
			Ours(Concate)& $\textbf{0.79}$& 24.17& $\textbf{0.48}$& $\textbf{3.85}$& $\textbf{0.09}$&$\textbf{0.86}$\\
			\bottomrule
			(c) LPCR& & & & & &\\
			w/o softmax threshold norm& 0.99& 31.14& 0.43& $\textbf{3.85}$& 0.11&1.29\\
			with positive class weight& 1.26& 30.30& $\textbf{0.48}$& 19.56& 0.12&1.39\\
			w/o hierarchical loss& 0.92& 28.44& 0.43& 10.38& 0.13&1.65\\
			Ours(full)& $\textbf{0.79}$& $\textbf{24.17}$& $\textbf{0.48}$& $\textbf{3.85}$& $\textbf{0.09}$&$\textbf{0.86}$\\
			\bottomrule
			(d) Loss Type& & & & & &\\
			L1 (predicting noise)& 3.82& 76.17& 0.37& 69.49& 0.17&3.08\\
			Huber (predicting noise)&1.24 &42.13 &0.47 &17.99 &0.14 &1.94\\
			L2 (predicting $f_0$)&1.51 &41.56 &0.44 &20.83 &0.14 &2.13\\
			Ours (L2, predicting noise)& $\textbf{0.79}$ &$\textbf{24.17}$ &$\textbf{0.48}$ &$\textbf{3.85}$ &$\textbf{0.09}$ &$\textbf{0.86}$ \\
			\midrule
			(e) Sample Steps& & & & & &\\
			32& 0.79& 25.62& $\textbf{0.49}$& 7.25& 0.11&1.17
			\\
			64&$\textbf{0.76}$&25.71&$\textbf{0.49}$&5.69&0.10&1.12\\
			Ours(128)& 0.79& $\textbf{24.17}$& 0.48& $\textbf{3.85}$& $\textbf{0.09}$&$\textbf{0.86}$\\
			\bottomrule
		\end{tabular}
	}
	\caption{Ablation study on GPAL dataset.}
	\label{tab:ablation}
	\vspace{-18pt}
\end{table}

\textbf{Results}: 
Examples of ground truth LiDAR and radar point clouds generated by the proposed and baseline methods are shown in Fig.~\ref{fig4}. The results reveal that the proposed method surpasses the baseline methods by generating denser and more accurate point clouds with fewer clutter points. Structural features of the scenes such as straight walls, curbs, and various objects in the Outdoor scene are also accurately predicted, which the baseline methods fail to do. In addition, these objects are highly unstructured. The quantitative results are listed in Tab.~\ref{tab:Quantitative}. These results convincingly demonstrate the superiority of our method in generating high-quality mmWave radar point clouds.

\subsubsection{2D Metric}   

\label{sec:addsuper}
Our evaluation metrics include CD, MHD, UCD, UMHD and FID, commonly used in previous works \cite{luan2024diffusion}.
As detailed in Sec.~\ref{sec:dataprocess}, the VoD dataset includes a forward-facing radar with a field of view (FOV) of 120° and a 360° LiDAR sensor. We evaluated our proposed method against state-of-the-art approaches, such as RadarHD \cite{16_prabhakara2023high} and R2DM \cite{82_nakashima2024lidar} in this dataset. Since the RadarHD method is specifically designed for 2D radar data, our method, which operates on 3D point clouds, is not directly applicable. To address this, we make the following adjustments for training on the VoD dataset: the 3D point clouds are projected into a bird's-eye view (BEV) image, where each pixel is assigned a value of either 0 or 255 to indicate the absence or presence of a point, respectively. Similarly, the R2DM method is adapted to use BEV images instead of range images, given the differing field of view (FOV) of the radar and LiDAR sensors. For our approach, the generated 3D point clouds are re-projected into 2D BEV images to facilitate comparison with the other baseline methods. Qualitative results are shown in Fig.~\ref{sub:fig3}.

\textbf{Results:}
For a fair comparison, we conduct all metrics in 2D, using only coordinates (x, y), the quantitative results are listed in the Tab.~\ref{tab:quantative_self_1}. On the VoD dataset, our method outperforms RadarHD~\cite{16_prabhakara2023high} and R2DM~\cite{82_nakashima2024lidar} across all evaluation metrics. Additionally, we achieve new state-of-the-art results in CD, UCD, and UMHD. This improvement can be attributed to our approach's ability to more accurately replicate the information in LiDAR point clouds. However, compared to~\cite{luan2024diffusion}, our method shows a slight decline in MHD and FID metrics. This is mainly due to our method working with 3D data, generating denser point clouds that may contain information not present in the original LiDAR point clouds. As a result, the enhanced point clouds no longer have a true correspondence with the LiDAR point clouds, leading to larger errors in the CD and MHD measurements.

\vspace{-2mm}
\subsection{Performance on Downstream Task: Registration}

Our enhanced point clouds present a precise overall layout while possessing enriched details, making them capable of downstream tasks. In this experiment, we demonstrate the capability of our enhanced point cloud for registration tasks. We adopt the setup in~\cite{luan2024diffusion} to evaluate our method on the test sets of the VoD data set. Point-cloud pairs with ground truth pose distances more than 1.5 m are chosen as test samples.

\textbf{Metrics:}  %
Following Bai's work \cite{bai2021pointdsc},  we use three evaluation metrics, namely (1) Registration Recall (RR), the percentage of successful alignment whose rotation error and translation error are below some thresholds, (2) Rotation Error (RE), and (3) Translation Error (TE). RE and TE are defined as 
\vspace{-2mm}
\begin{eqnarray}
	RE(\hat{R} )=acos\frac{Tr(\hat{R^{T}}R^{*}  )-1}{2} , TE(\hat{t})={\left \|\hat{t}-t^{*}    \right \| }_{2}, 
	\vspace{-2mm}
\end{eqnarray}
where $R^*$ and $t^*$ denote the ground-truth rotation and translation, respectively. For RR, a registration result is considered successful if \( TE < 0.5\,\text{m} \) and \( RE < 5^\circ \).

\textbf{Results:}%
In Tab.~\ref{tab:Regis}, we present registration results on raw radar point clouds and enhanced radar point clouds generated by our Radar-diffusion utilizing the classical registration method, PointToPoint-ICP. We directly apply our method to various point cloud datasets to assess the registration performance. As demonstrated, the enhanced point clouds exhibit high consistency and accuracy, ensuring reliable registration. Moreover, the additional detailed information provided by the enhanced point clouds enables a more robust registration compared to radar point clouds.

\vspace{-2mm}
\subsection{Performance on Downstream Task: Object Detection}
In this experiment, we demonstrate the capability of our enhanced point cloud for object detection tasks. We consider object detection performance on three object classes: car, pedestrian and cyclist. 

\textbf{Metrics:} We use the performance measure: Average Precision (AP). We calculate the intersection over union (IoU) of the predicted and ground truth bounding boxes in 3D, and require a $50\%$ overlap for car, and $25\%$ overlap for pedestrian and cyclist classes. Mean AP (mAP) is calculated by averaging class-wise results. Following \cite{palffy2022multi}, we report results for two regions: the entire annotated region and Driving Corridor.

\textbf{Results:} Tab.~\ref{table:flyingthing3d} reports the comparison results, where the enhanced point clouds produced by our method outperform the raw point clouds in the PointPillar-based object detection method across all categories and evaluation metrics. Specifically, the mAP values for two scenes are improved by 5.45\% and 10.97\%, respectively, and the detection success rate for cyclists in the Driving Corridor is significantly increased by 24.88\%. We further provide the visual comparisons in Fig.~\ref{sub:fig::od}, where we can see that: our method generates more accurate 3D bounding boxes, which are consistent with the ground truth boxes, see the green boxes in the first row.

\begin{table}[H]
	\centering
	\resizebox{0.86\linewidth}{!}{ 
		\begin{tabular}{@{}l c  @{\hskip 10pt}c cc}
			\toprule
			& & & $[succ./all]$&$[succ./all]$\\
			Case& & RR(\%) $\uparrow$& RE($^\circ$) $\downarrow$&TE(m)$\downarrow$\\
			\midrule
			1scan& Raw& 25.6& 1.39/2.18& 0.30/1.29\\
			& Ours& 57.3& 1.11/1.86& 0.22/0.96\\
			3scans& Raw& 70.7& 1.15/1.36& 0.25/0.74\\
			& Ours& 72.0& 1.08/1.39& 0.19/0.60\\
			5scans& Raw& 85.7& 0.79/$\textbf{0.84}$& 0.18/0.37\\
			& Ours& $\textbf{89.0}$& $\textbf{0.74/}$0.86& $\textbf{0.17/0.28}$\\
			\bottomrule
		\end{tabular}
	}
	\caption{Ablation study for registration performance using enhanced point cloud in VoD dataset.}
	\label{tab:Registration}
	\vspace{-14pt}
\end{table}

\subsection{Ablation Study}
\vspace{-1mm}
We conduct a series of experiments to evaluate the key components of our R2LDM framework, as detailed in Tab.~\ref{tab:ablation}. (a) We investigated the impact of different diffusion model architectures on the model's performance. The UNet-based diffusion model outperformed other architectures, such as transformer-based and MLP-based models, within our framework. (b) We explored encoding conditions through cross-attention mechanisms, but the results were inferior to the concatenation approach. Additionally, we investigated the inclusion of extra conditions such as geometry features and intensity, but these did not yield significant improvements. (c) We validated the effectiveness of the redesigned LPCR module, including the softmax threshold normalization and hierarchical loss methods, in enhancing model performance. Furthermore, we demonstrate that the application of positive class weighting is not suitable for this task. Moreover, in (d) and (e), we investigated the impact of different loss functions and the number of sampling steps during inference on model performance. Ultimately, we determined that the default loss function should be L2, directly predicting the noise, with 128 sampling steps, as this configuration yielded the most balanced performance. In Tab.~\ref{tab:Registration}, we present the performance of the raw and enhanced point clouds in the point cloud registration task for various input frames. Notably, the enhanced point clouds from a single scan of radar point clouds showed the greatest improvement in recall, while merging five scans of radar point clouds resulted in the best overall performance.

\section{Conclusion}
In this work, we present R2LDM, a novel framework for 4D radar point cloud super-resolution, enhancing autonomous perception in adverse weather. R2LDM employs a latent voxel-based diffusion model to unify radar and LiDAR in a shared latent space, effectively capturing 3D structures for improved reconstruction. Based on this, the latent point cloud reconstruction module generates dense, LiDAR-like point clouds while preserving fine details. A two-stage training strategy ensures robust reconstruction across diverse environments. Experiments show that R2LDM achieves state-of-the-art performance, significantly improving point cloud registration and object detection, thereby establishing it as a reliable solution for radar-based perception, enabling all-weather operation in complex scenarios. In future work, we will explore how to retain key radar-specific properties, such as velocity and RCS, in enhanced point clouds.


\addtolength{\textheight}{-12cm}

%
%


\bibliographystyle{IEEEtran}  
\bibliography{ref}

\end{document}